\documentclass[letterpaper, 10 pt, conference]{ieeeconf}  

\IEEEoverridecommandlockouts                              

\usepackage{cite}

\usepackage{graphicx,color}  
\usepackage{amsfonts}  
\usepackage{amsmath}
\usepackage{amssymb}

\graphicspath{ {images/} }

\title{\LARGE \bf
Frontal Plane Bipedal Zero Dynamics Control$^*$
}

\author{Arthur Castello Branco de Oliveira$^{1}$, Guilherme Scabin Vicinansa$^{2}$, \\ Paulo S\'ergio Pereira da Silva$^{3}$, Bruno Augusto Ang\'elico$^{4}$
	\thanks{*The authors would like to thanks Coordena\c c\~ao de Aperfei\c coamento de Pessoa de N\'ivel Superior (CAPES) for the grants that founded this research}
	\thanks{$^{1}$Arthur C. B. Oliveira is with Escola Polit\'ecnica,
		University of S\~ao Paulo, AV Prof. Luciano Gualberto 158 - S\~ao Paulo/SP, Brazil
		{\tt\small arthur.castello.oliveira@usp.br}}%
	\thanks{$^{2}$Guilherme S. Vicinansa is with the Coordinated Science Laboratory at the University of Illinois at Urbana-Champaign,
		Urbana, Illinois, USA
		{\tt\small gs16@illinois.edu}}%
	\thanks{$^{3}$Paulo S. P. da Silva is with Escola Polit\'ecnica,
	University of S\~ao Paulo, AV Prof. Luciano Gualberto 158 - S\~ao Paulo/SP, Brazil
	{\tt\small get.email@usp.br}}%
	\thanks{$^{4}$Bruno A. Ang\'elico is with Escola Polit\'ecnica,
	University of S\~ao Paulo, AV Prof. Luciano Gualberto 158 - S\~ao Paulo/SP, Brazil
	{\tt\small angelico@usp.br}}%
}

\begin{document}

\maketitle
\thispagestyle{empty}
\pagestyle{empty}

\begin{abstract}

In bipedal gait design literature, one of the common ways of generating stable 3D walking gait is by designing the frontal and sagittal controllers as decoupled dynamics. The study of the decoupled frontal dynamics is, however, still understudied if compared with the sagittal dynamics. In this paper it is presented a formal approach to the problem of frontal dynamics stabilization by extending the hybrid zero dynamics framework to deal with the frontal gait design problem.

\end{abstract}

\section{Introduction}

When extending sagittal gait designs to 3D bipedal robots it is common to design the frontal dynamics as decoupled from the sagittal one. In \cite{pratt1999exploiting} \cite{da20162d} and \cite{luo2012planning} the authors design a decoupled frontal controller using foot placement to contain the post-impact kinetic energy of the frontal dynamics within the potential energy barrier and generate stable walking.

Even so, the frontal dynamics is still treated as secondary in the literature. While it is easy to find papers that study, and propose gait designs for the decoupled sagittal dynamics, \cite{apostolopoulos2016online} \cite{apostolopoulos2015settling} \cite{chevallereau2008stable} \cite{dai2012optimizing} \cite{ames2014rapidly} \cite{galloway2015torque} etc, the literature is still scarce in the study of the decoupled frontal dynamics. This is likely due to the fact that the frontal dynamics has been successfully stabilized so far with simple foot placement control techniques.

In this paper a more formal approach to the frontal dynamics stabilization is proposed by extending the hybrid zero dynamics (HZD) framework to the frontal dynamics. The HZD is a formalistic approach to sagittal gait design method that does not consider any simplification on the design model. It allows for easily specification of style constraints while obtaining an energy efficient  gait. There is also works in the literature that extend this method to deal with terrains irregularities, \cite{apostolopoulos2016online}, and online changes in the gait style specification \cite{westervelt2007feedback} and \cite{apostolopoulos2015settling}.

As a side note, it is important to note that extending planar gaits is not the only method proposed in the literature to generate 3D stable gaits. In \cite{shih2007asymptotically} and \cite{chevallereau2009asymptotically} the authors extender the HZD technique for 3D robots and generated a stable 3D gait, and in \cite{hereid20163d} the author used virtual constraints and nonlinear programming to also obtain a 3D stable gait.

If considering other techniques as well, there are other examples os successful 3D gait design, as in \cite{pratt2006capture} \cite{koolen2012capturability} and \cite{pratt2012capturability}, where the authors explore the use of capturability to generate stable gait.

The design of planar gaits is, however, simpler and less computationally consuming than the design of 3D gaits, which motivates the study of ways to generate stable 3D walking gaits from decoupled planar techniques.

The main goal of this paper is to contribute to the generation of stable 3D gaits from decoupled planar ones by proposing a technique for the frontal dynamics that allow fine tuning of the gait style by specification of nonlinear constraints for the optimization process. The actual generation of a 3D stable gait is not addressed in this paper but some problems, related to the potential energy barrier, that may arise when trying to compose the decoupled gaits into a 3D one are foreshadowed in the results of this paper.

\section{Modeling and Description of the System}
\label{sec:model}

In this section the considered model and related hypothesis are presented aiming to highlight both the similarities and differences between the frontal model and the typical sagittal model found in the literature.

\subsection{Robot Hypothesis}

This paper adopts the usual assumptions widely used in the bipedal locomotion literature, particularly the hypotheses presented in \cite{westervelt2007feedback} section $10.2.1$ were adapted to best suit this application as shown bellow (the same notation was kept for consistency).

The robot is assumed to:
\begin{itemize}
	\item[HR1] be comprised of $N$ rigid links connected by $(N-1)$ ideal revolute joints to form a single open kinematic chain;
	\item[HR2] be planar, with motion restricted to the frontal plane;
	\item[HR3] be bipedal, with two symmetric legs connected at opposite ends of a common joint with nonzero length, called the hip. Both legs are terminated in symmetrical feet of nonzero length;
	\item[HR4] be independently actuated at each of the $(N-1)$ ideal revolute joints;
	\item[HR5] have two finite symmetrical feet with zero height at the ankle in relation to the contact surface with the ground.
\end{itemize}

Further hypotheses were made about the desired gait for the frontal dynamics.
\begin{itemize}
	\item[HG1] Walking consists of four continuous phases: one during which the ankle rotational speed is strictly positive (the rising phase), one when its speed is strictly negative (the falling phase), the double support phase and the reversal phase, when the ankle speed changes direction;
	\item[HG2] During all the phases the stance foot remains flat on the ground and does not slip;
	\item[HG3] During the gait the angular moment about the stance ankle is instantaneously zero during the reversal phase --- transition from rising to falling phases;
	\item[HG4] The reversion occurs only once for each step. It is instantaneous, and, in steady state, it always happens for the same value of ankle angular position;
	\item[HG5] The double support phase is instantaneous and the associated impact can be modeled as a rigid contact;
	\item[HG6] The positions and velocities are continuous across both the transitions;
	\item[HG7] In steady state, the beginning and end position of the swing leg, in each step, are strictly the same resulting in no movement in the frontal plane;
	\item[HG8] In steady state the motion is symmetric with respect to the two legs.
\end{itemize} 

The impact hypotheses (HI1 to HI7) are the same or extremely alike and therefore were omitted here.

\subsection{Angle Conventions}

\begin{figure}[h!]
	\centering
	\includegraphics[scale=0.4]{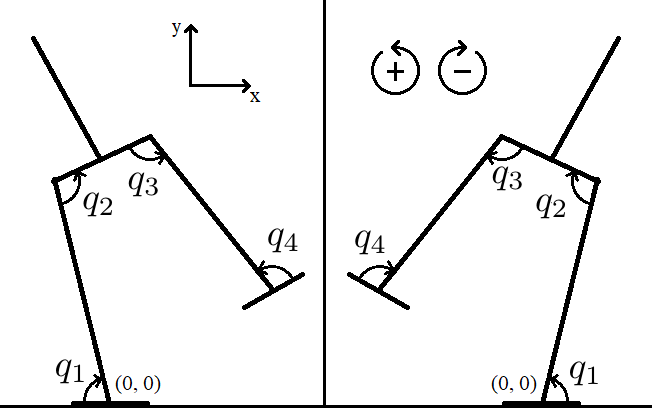}
	\caption{Modeling conventions for when the swing foot is the left one or the right one \label{fig:AngleFrontal}}
\end{figure}

The angle conventions for the frontal dynamics are shown in Fig. \ref{fig:AngleFrontal}. 

Note that, although the dynamics differ whether the left or the right foot are the stance one, from the chosen angles perspective the joint torques are the same except for a minus signal. For this reason, hereinafter it will be considered that the stance leg is the left one, without loss of generality\footnote{If the right foot is the stance foot, it is only necessary to invert the angles when computing the control torques and then invert the required torques before applying it to the system}.

The parameters used for this model are presented in table \ref{tbl:parammodel}, with the link's inertias taken in reference to the center of mass (CoM), and the link's axes positioned so that the origin is at the joint with the previous link, $x_i$ is positive in the direction to the next joint and $y$ is oriented so that $z$ enters the plane and the base is positive.%
\begin{table}
	\small
	\caption{System's Parameters\label{tbl:parammodel}}
	\begin{center}
	\begin{tabular}{c|c|c}
		\hline
		1st link CoM & \{0.16,  0.00\} & [$m$] \\
		2nd link CoM & \{0.10, -0.20\} & [$m$] \\
		3rd link CoM & \{0.16,  0.00\} & [$m$] \\
		4th link CoM & \{0.00,  0.00\} & [$m$] \\
		1st link Inertia & 0.400 & [$kg \ m^2$] \\
		2nd link Inertia & 5.530 & [$kg \ m^2$] \\
		3rd link Inertia & 0.400 & [$kg \ m^2$] \\
		4th link Inertia & 0.030 & [$kg \ m^2$] \\
		1st link Mass & 12.15 & [$kg$] \\
		2nd link Mass & 36.00 & [$kg$] \\
		3rd link Mass & 12.15 & [$kg$] \\
		4th link Mass & 0.200 & [$kg$] \\
		1st link Length & 0.800 & [$m$] \\
		2nd link Length & 0.200 & [$m$] \\
		3rd link Length & 0.800 & [$m$] \\
		4th link Length & 0.100 & [$m$] 
	\end{tabular}
	\end{center}
\end{table}%
\subsection{Rising and Falling Models}

During both the rising and falling phases the stance foot is supposed pinned to the ground and thus a 4-d.o.f. model is enough to describe the system dynamics. Using the method of Lagrange yields the model presented in Eq. (\ref{eq:pinnedmodel}), where $q = [q_1 \ q_2 \ q_3 \ q_4]^\top \in \mathcal{Q}$ is the joints positions vector, $u_b = [u_2 \ u_3 \ u_4]^\top \in \mathcal{U}_b$ is the control input vector, and $u_1\in\mathbb{R}$ is the ankle torque.%
\begin{equation}
	\label{eq:pinnedmodel}
	D_s(q)\ddot{q}+C_s(q, \dot{q})+G_s(q)=\underbrace{\begin{bmatrix}
	0_{1\times3} \\ I_{3\times3}
	\end{bmatrix}}_{B_s}u_b+\begin{bmatrix}
	1 \\ 0_{3\times1}\end{bmatrix}u_1
\end{equation}

The difference between the rising and falling models is due to the virtual constraints parameters, and consequently in the control signal $u_b$ and $u_1$.

The system can then be described in a state space form, as presented in Eq. (\ref{eq:sspinned}), with $x = [q \ ; \ \dot{q}]$.%
\begin{equation}	
	\label{eq:sspinned}
	\dot{x} = \underbrace{\begin{bmatrix}
		\dot{q} \\ D_s^{-1}(-C-G+\begin{bmatrix}
		1 \\ 0_{3\times1}\end{bmatrix}u_1)
	\end{bmatrix}}_{f_s(x, u_1)}
	+
	\underbrace{\begin{bmatrix}
		0_{4\times 4} \\
		D_s^{-1}B_s
	\end{bmatrix}}_{g_s(x)}u_b
\end{equation}

Note that to describe the internal dynamics and avoid high ankle torques that would result in foot rotation, the ankle torque is not considered in the virtual constraints design. This extra degree of actuation is later used to enforce a desired position of the foot rotation indicator (FRI, as defined in \cite{goswami1999postural}) in a similar manner as is described in \cite{westervelt2007feedback}, ch. 11.

\subsection{Impact Model}

The impact, or the transition from falling to rising, is modeled exactly as in \cite{westervelt2007feedback}. The impact equation is shown in Eq. (\ref{eq:impactModel}), where $x_f^-$ is the state of the falling phase just before transition and $x_r^+$ is the state of the rising phase just after transition.%
\begin{eqnarray}
	\label{eq:impactModel}
	x_r^+ = \Delta x_f^- 
	& \forall x_f^-\in\mathcal{S}_f^r
\end{eqnarray}

In Eq. (\ref{eq:impactModel}), $\Delta$ can be calculated using the unpinned 6-d.o.f. model of this system, as described in \cite{westervelt2007feedback}.

\subsection{Hybrid Model Description}

The hybrid model used in this paper can be expressed as a nonlinear hybrid system described in two different manifolds for the rising and falling phases, as in Eq. (\ref{eq:hybridsystem}), where $p_4^v(q)$ is the vertical position of each of the swing foot ends for the configuration $q$.%
\begin{equation}
	\small
	\label{eq:hybridsystem}
	\begin{split}
		\Sigma_r& = \begin{cases}
			\mathcal{X}_r &= T\mathcal{Q} \\
			\mathcal{F}_r: \dot{x_r} &= f_s(x_r, u_{1r})+g_s(x_r)u_{br} \\
			\mathcal{S}_r^f &= \{x_r^-\in\mathcal{X}_r \ | \ \dot{q}_{1r}^- = 0, \ \ddot{q}_{1r}^-<0\} \\
			\mathcal{T}_r^f: x_f^+ &= x_r^- 
		\end{cases} \\
		\Sigma_r& = \begin{cases}
			\mathcal{X}_f &= T\mathcal{Q} \\
			\mathcal{F}_f: \dot{x_f} &= f_s(x_f, u_{1f})+g_s(x_f)u_{bf} \\
			\mathcal{S}_f^r &= \{x_f^-\in\mathcal{X}_f \ | \ p_{4}^v(q_f^-) = 0_{2\times1} \} \\
			\mathcal{T}_f^r: x_r^+ &= \Delta x_f^- 
		\end{cases}
	\end{split}
\end{equation}

In Eq. (\ref{eq:hybridsystem}), $\mathcal{X}_{r,f}$ are the state manifold of each phase, $\mathcal{F}_{r,f}$ are the dynamics on each manifold, $\mathcal{S}_{r,f}^{f,r}$ are the switching sets, and $\mathcal{T}_{r,f}^{f,r}$ are the transition functions applied to its respective switching set.

\section{Zero Dynamics}
\label{sec:Zdyn}

\subsection{Rising and Falling Zero Dynamics} 

Similarly to the sagittal robot with nontrivial feet, the system's zero dynamics exists for a set of outputs $y = h(q) = q_b-h_d(\beta(q))$, where $\beta$ is such that $[\beta ; h]$ is a diffeomorphism defined in $\mathcal{Q}$ to its image, and $h_d(\beta)$ is the vector of desired joints positions, parametrized by $\beta$. Particularly, $\beta = q_1$ satisfies this condition and allows the computation of the zero dynamics. This is thoroughly described in \cite{westervelt2007feedback}, ch.11, and only the results are shown here.

Consider that both $q_b^d$ and $p_{FRI}^{h,d}$ (i.e. the desired horizontal FRI position) are expressed as a set of Bezi\'er polynomials, $h_d$ and $h_d^{FRI}$, with $b$ and $b_{FRI}$ being its parameters matrix. Consider also that $u_b$ and $u_1$ are chosen, respectively, so that the resulting zero dynamics is forward invariant and that the desired FRI position is tracked. The resulting zero dynamics for both rising and falling phases (with the difference between them lying in the values of $b$ and $b_{FRI}$ at each phase) are presented in Eqs. (\ref{eq:zerodyn1}) and (\ref{eq:zerodyn2}), where $\sigma_{FRI}$ is the system's momentum in relation to the FRI point.%
\begin{eqnarray}
	\dot{\beta}=&\kappa_1^{FRI}(\beta)\sigma_{FRI} \label{eq:zerodyn1} \\
	\dot{\sigma}_{FRI}=&\kappa_2^{FRI}(\beta)+\kappa_3^{FRI}(\beta)\sigma_{FRI}^2 \label{eq:zerodyn2}
\end{eqnarray}

Dividing Eqs. (\ref{eq:zerodyn1}) and (\ref{eq:zerodyn2}), Eq. (\ref{eq:edo1}) is obtained. Performing, then, the variable change $\frac{\sigma_{FRI}^2}{2} = \zeta_{FRI}$ results in Eq. (\ref{eq:edo2}).%
\begin{eqnarray} 
	\label{eq:edo1}
	\frac{d\sigma_{FRI}}{d\beta} =& \frac{\kappa_2^{FRI}(\beta)}{\kappa_1^{FRI}(\beta)\sigma_{FRI}}+\frac{\kappa_3^{FRI}(\beta)}{\kappa_1^{FRI}(\beta)}\sigma_{FRI} \\
	\label{eq:edo2}
	\frac{d\zeta_{FRI}}{d\beta} =& \frac{\kappa_2^{FRI}(\beta)}{\kappa_1^{FRI}(\beta)}+2\frac{\kappa_3^{FRI}(\beta)}{\kappa_1^{FRI}(\beta)}\zeta_{FRI}
\end{eqnarray}

Note that, a priori, this is only true for $\dot{\beta}\neq0$ due to the fact that Eq. (\ref{eq:edo1}) has a singularity for $\sigma_{FRI}=0$. This represents a problem for the robot's hypothesis, since it is supposed that a speed reversion occurs during operation. This particular point is treated in the next subsection and ignored for the continuous phases.

Eq. (\ref{eq:edo2}) is a differential equation, linear in $\zeta_{FRI}$ and with $\beta$-varying parameters whose solution is presented in Eq. (\ref{eq:soledo}).%
\begin{equation}
\small
\begin{split}
	\label{eq:soledo}
	\zeta_{FRI} &= \delta_{FRI}(\beta)^2\zeta_{FRI}^+-V_{\mathcal{Z}}^{FRI}(\beta) \\
	\delta_{FRI} &= exp\bigg({\int_{\beta^+}^{\beta}\frac{\kappa_3^{FRI}(\tau)}{\kappa_1^{FRI}(\tau)}d\tau}\bigg)	\\
	V_{\mathcal{Z}}^{FRI} &= - \int_{\beta^+}^{\beta}exp\bigg(2{\int_{\tau_2}^{\beta}\frac{\kappa_3^{FRI}(\tau_1)}{\kappa_1^{FRI}(\tau_1)}d\tau_1}\bigg)\frac{\kappa_2^{FRI}(\tau_2)}{\kappa_1^{FRI}(\tau_2)}d\tau_2
\end{split}
\end{equation}

This equation describes the system restricted to its zero dynamics and in relation to the FRI point. It is convenient, however, to determine the zero dynamics description in relation to the stance foot ankle, being $\sigma$ the momentum of the system in relation to the stance ankle.

\subsection{Impact and Reversal Zero Dynamics}

The restriction of the impact to the zero dynamics is completely analogue to the usually presented in the literature, as in \cite{westervelt2007feedback}. For the case of a robot with nontrivial feet the impact can be expressed in terms of the change in angular momentum at the stance ankle, as expressed in Eq. (\ref{eq:restimpact}), where the underscript $4$ indicates the fourth link (that is, the swing foot), the underscripts $f$ and $r$ indicate the falling and rising phases respectively, and the undescript $cm$ indicates that the variable is about the center of mass. The variable $J_i$ is the inertial moment of link $i$ in relation to its center of mass, $m_i$ is the link's mass and $\omega_i$ is the link's absolute angular velocity.%
\begin{equation}
	\label{eq:restimpact}
	\begin{split}
		\sigma_r^+ = \sigma_f^- -p_{4f}^{h-}p_{cmf}^{v-}-J_{4}\omega_{4f}^- \\ -m_{4}p_{4cmf}^{h}\dot{p}_{4cmf}^{v-}+m_{4}p_{4cm}^{v}\dot{p}_{4cmf}^{h-}
	\end{split}
\end{equation}

When restricting Eq. (\ref{eq:restimpact}) to the zero dynamics, the terms become linearly related to $\sigma_f^-$, allowing it to be put in evidence, resulting in Eq. (\ref{eq:restimpactsimp}), where $\delta_{z}(\beta)$ is function only of $\beta$, and $\beta_f^-$ is the value of $\beta$ just before the transition from falling to rising.%
\begin{equation}
	\label{eq:restimpactsimp}
	\sigma_r^+ = \delta_{z}(\beta_f^-) \ \sigma_f^-
\end{equation}

As for the reversal, the transition is smooth as stated in Eq. (\ref{eq:invtran}).%
\begin{equation}
	\label{eq:invtran}
	\sigma_f^+ = \sigma_r^-
\end{equation}

However, since the system is now described in its zero dynamics, the singularity must be analyzed. 

Being $\beta_r^+$ and $\beta_r^-$ the values of the angle $\beta$ at the beginning and at the end of the rising phase. For the full system, it is known that $\dot{\beta} = 0$ is equivalent to $\sigma_{FRI} = 0$, that is equivalent to $\zeta_{FRI} = 0$. Indeed, from Eq. (\ref{eq:soledo}), for a given value of $\beta = \beta_r^-$, it is possible that%
$$ \zeta_{FRI}(\beta_r^-) = \delta_{FRI}(\beta_r^-)^2\zeta_{FRI}^+-V_{\mathcal{Z}}^{FRI}(\beta_r^-) = 0$$

\noindent meaning that, despite Eq. (\ref{eq:edo1}) having a singularity for $\sigma_{FRI} = 0$, Eq. (\ref{eq:edo2}) allows the computation of its solution for $\zeta_{FRI}(\beta_r^-) = 0 \iff \sigma_{FRI}(\beta_r^-)=0$. Due to the continuity of $\sigma_{FRI}(\beta)$ and $\zeta_{FRI}(\beta)$ with relation to $\beta$, since%
\begin{equation*}
\begin{matrix}
	\frac{\sigma_{FRI}(\beta)^2}{2} = \zeta_{FRI}(\beta), &\beta\in[\beta_r^+, \ \beta_r^-[
\end{matrix}
\end{equation*}

\noindent then%
\begin{equation*}
	\lim\limits_{\beta\rightarrow\beta_r^-}\frac{\sigma_{FRI}(\beta)^2}{2} = \lim\limits_{\beta\rightarrow\beta_r^-}\zeta_{FRI}(\beta) = \zeta_{FRI}(\beta_r^-) = 0
\end{equation*}

\noindent thus,%
\begin{equation*}
	\lim\limits_{\beta\rightarrow\beta_r^-}\sigma_{FRI}(\beta) = 0 \triangleq \sigma_{FRI}(\beta_r^-)
\end{equation*}

\noindent which is consistent with the expected result. Therefore, the reversal is well defined for the restricted system. A point $\beta_r^-$ is, then, called an reversal point of the hybrid system if at it:

\begin{itemize}
	\item[HIP1] the system is halted, that is, $\dot{\beta}(\beta_r^-) = 0 \iff \sigma_{FRI}(\beta_r^-) = 0 \iff \zeta_{FRI}(\beta_r^-) = 0$;
	\item[HIP2] the ankle rotational speed changes direction, that is $\dot{\sigma}(\beta_r^-) < 0 \iff \ddot{\beta}(\beta_r^-)<0$ (considering the left foot on the ground and using the chosen angle and axis conventions).
\end{itemize}

\subsection{Hybrid Poincar\'e Function}

With Eq. (\ref{eq:soledo}) it is possible to calculate the values of $\zeta_{FRI}$ at the end of each continuous phase, given their value at the beginning and the values of $\beta$ at each transition, that is,%
\begin{equation}
	\zeta_{FRIr,f}^- = \delta_{FRI}(\beta_{r,f}^-)^2 \ \zeta_{FRIr,f}^+-V_z^{FRI}(\beta_{r,f}^-)
\end{equation}.

Describing this equation in relation to the stance ankle instead of the FRI point, considering $\zeta = \frac{\sigma^2}{2}$, results in%
\begin{eqnarray}
\delta_a^{FRI}(\beta_{r,f}^-)^2 \ \zeta_{r,f}^- = \delta_{FRI}(\beta_{r,f}^-)^2 \ \delta_a^{FRI}(\beta_{r,f}^+)^2 \ \zeta_{r,f}^+ -\nonumber\\ -V_z^{FRI}(\beta_{r,f}^-) \\
\zeta_{r,f}^- = \underbrace{\delta_{FRI}^{a}(\beta_{r,f}^-)^2 \ \delta_{FRI}(\beta_{r,f}^-)^2 \ \delta_a^{FRI}(\beta_{r,f}^+)^2}_{\delta_{r,f}^2} \ \zeta_{r,f}^+ -\nonumber \\ -\delta_{FRI}^{a}(\beta_{r,f}^-)^2 \ V_z^{FRI}(\beta_{r,f}^-)  
\end{eqnarray}

\noindent where $\delta_{FRI}^{a} = (\delta_{a}^{FRI})^{-1}$ are the conversion from the FRI to the ankle and from the ankle to the FRI, respectively, being easily calculated for a given $\beta$ using the angular momentum transfer theorem, as shown in \cite{westervelt2007feedback}, ch. 11.

Furthermore, with Eqs. (\ref{eq:restimpactsimp}) and (\ref{eq:invtran}) it is possible to calculate $\zeta_{FRI}$ at the beginning of each phase, knowing its value at the end of the previous phase, that is,%
\begin{eqnarray}
	\label{eq:pcr}
	\rho_r(\zeta_f^-) \triangleq \zeta_{r}^- = \delta_{r}^2 \ \delta_z^2 \ \zeta_{f}^- -\delta_{FRI}^{a}(\beta_{r}^-)^2 \ V_{zr}^{FRI}(\beta_{r}^-) \\
	\label{eq:pcf}
	\rho_f(\zeta_r^-) \triangleq \zeta_{f}^- = \delta_{f}^2 \ \zeta_{r}^- -\delta_{FRI}^{a}(\beta_{f}^-)^2 \ V_{zf}^{FRI}(\beta_{f}^-)
\end{eqnarray}

Finally composing Eqs. (\ref{eq:pcr}) and (\ref{eq:pcf}) it is possible to obtain the hybrid Poincar\'e function shown in Eq. (\ref{eq:pcc}) and with domain of definition $\mathcal{D} := \{\zeta_f^->0 \ | \ \zeta_{f,r}(\zeta_f^-)\geq 0, \ and \ \exists! \beta_{r}^-\rightarrow \dot{\beta}(\beta_{r}^-) = 0, \ \ddot{\beta}(\beta_{r}^-) < 0\}$.%
\begin{equation}
	\label{eq:pcc}
	\begin{split}
		&\rho(\zeta_f^-) = \rho_f\circ\rho_r(\zeta_f^-) = \\
		\delta_f^2 \ \delta_r^2 \ \delta_z^2 \ \zeta_f^- -&\delta_f^2 \ \delta_{FRI}^a(\beta_r^-)^2 \ V_{zr}^{FRI}(\beta_r^-) -\\ -&\delta_{FRI}^{a}(\beta_f^-)^2 \ V_{zf}^{FRI}(\beta_f^-)
	\end{split}
\end{equation}

With this Poincar\'e function it is possible to find the closed form of the fixed point, such that $\rho(\zeta_f^*) = \zeta_f^*$, as presented in Eq. (\ref{eq:fixedp}).%
\begin{equation}
	\label{eq:fixedp}
	\zeta_f^* = - \frac{\delta_f^2 \ \delta_{FRI}^a(\beta_r^-)^2 \ V_{zr}^{FRI}(\beta_r^-)+\delta_{FRI}^a(\beta_f^-)^2 \ V_{zf}^{FRI}(\beta_f^-)}{1-\delta_f^2 \ \delta_r^2 \ \delta_z^2}
\end{equation}

Then, the limit cycle exists in the hybrid zero dynamics if $\zeta_f^*\in\mathcal{D}$, and is stable if $0<\delta_f^2 \ \delta_r^2 \ \delta_z^2<1$.

\subsection{Inversion Point Computation}

Although there is a closed form for the fixed point, it depends on the value of the reversal point. While the impact can be calculated {\it a priori}, since it depends only on the robot's configuration --- that is known a priori for the restricted system if the constraints are parametrized by a Bezi\'er polynomial --- the reversion depends mainly on the post impact velocity of the system and, therefore, can not be calculated with only the configuration at the beginning or end of each phase.

To calculate the reversal point, the marginal condition $\zeta_r^- = 0$ is applied to Eq. (\ref{eq:pcr}) together with the closed form of the fixed point ($\zeta_f^- = \zeta_f^*$), resulting in Eq. (\ref{eq:marginalcond}).%
\begin{equation}
	\label{eq:marginalcond}
	\begin{split}
		Z(\beta_r^-) \triangleq (1-\delta_f^2 \ \delta_r^2 \ \delta_z^2)\rho_r(\beta_r^-) = \\ =-\delta_r^2 \ \delta_z^2 \ \delta_{FRI}^a(\beta_f^-)^2 \ V_{zf}^{FRI}(\beta_f^-)-\\-\delta_{FRI}^a(\beta_r^-)^2 \ V_{zr}^{FRI}(\beta_r^-)
	\end{split}
\end{equation}

By definition $Z(\beta_r^-) = 0$ if $\beta_r^-$ is a reversal point. Furthermore, being $q_r^- = [\beta_r^- \ ; \ q_{br}^-]$, where $q_{br}^-$ is known from the Bezi\'er parameters, then if $p_{cm}^h(q_r^-) > 0$ and $Z(\beta_r^-) = 0$, $\beta_r^-$ is an reversal point.

Therefore, to find the reversal point one needs only to numerically find the zero of $Z(\beta)$. However, in order for the limit cycle in the zero dynamics to be also a limit cycle in the full model, the parameters need to be hybrid invariant. While the forward invariance can be easily enforced by the choice of feedback law, the transition invariance requires a changing in the first two columns of the Bezi\'er parameters matrix $b$, in accordance to the restrictions presented in \cite{westervelt2007feedback} (with little change for the reversion invariance, where there is no impact).

The problem is that, to compute the transition invariance restrictions for the parameters, the values of $\beta_r^-$ and $\beta_f^+$ (that are equal due to the reversion description) are necessary, and the invariant parameters are needed so that the computation of $Z(\beta)$, and consequently the computed transition point, is valid not only for the restricted system but for the complete system as well.

In this work, to solve this problem, an algorithm of successive approximations was used in order to find $\beta_r^-$, that is, the invariance is imposed on the parameters using an initial guess of $\bar{\beta}_r^-$, then the zero of $Z(\beta)$ was found for this set of parameters and the invariance was recalculated for the new value of $\beta_r^-$.

Although the conditions to convergence of this algorithm were not yet studied, the algorithm converged in less then 10 steps for every value of $b$ that had an unique reversal point (the parameters were given by the Matlab {\it fmincon} function). However, if the reversal point does not exists or is not unique, $Z(\beta)$ might not be defined for some points, which motivates the search of existence conditions for the reversal point.

\subsection{Existence of the Inversion Point}

Let $\beta_r^+$ be the value of $\beta$ at the beginning of the rising phase, $\beta_r^-$ the value at the ending of the rising phase, $\beta_f^+$ the value at the beginning of the falling phase and $\beta_f^-$ the value at the ending of the falling phase, with $\beta_r^+$ and $\beta_f^-$ known {\it a priori}. 

Since, by the measurement conventions of this paper $\beta<0$, suppose, without loss of generality, that $\beta_r^+>\beta_f^-$ (if not, the prof is analogous). If exists $\beta_r^-$ that satisfies HIP1 and HIP2, then $\beta_r^->\beta_r^+$. This is straightforward and the proof is not shown here, but can be easily concluded if it is considered that $\beta_r^-<\beta_r^+$ results in a fall after impact.

Furthermore, consider $q_{br}^- = q_{bf}^+$ the configuration variables right at the end of the rising phase and at the beginning of the falling phase, respectively, also consider $p_{FRIr}^{h,d-}$ the desired horizontal FRI at the end of the rising phase, and assuming that $p_{cm}^h([\beta_r^+ \ ; \ q_{br}^-])>p_{FRIr}^{h,d-}$. Considering $\beta_r^{max}$ so that,%
$$ p_{cm}^h([\beta_r^{max} \ ; \ q_{br}^-])=p_{FRIr}^{h,d-}, $$

\noindent then, if there exists $\beta_r^-$ that satisfies HIP1 and HIP2, $\beta_r^-\in]\beta_r^+, \ \beta_r^{max}[$.

To prove that, suppose that $\beta_r^->\beta_r^{max}$ is an reversal point, {\it i.e.}, that satisfies HIP1 and HIP2. Then, using Eq. (\ref{eq:zerodyn2}), it follows that:%
$$\dot{\sigma}_{FRI}^- = \kappa_2^{FRI-}+\kappa_3^{FRI-} \ \underbrace{\sigma_{FRI}^-}_{=0}.$$

Using the expression for $\kappa_2^{FRI}$ calculated in \cite{westervelt2007feedback}, it results in%
$$\dot{\sigma}_{FRI}^- = -m_{tot} \ g_0 \ (p_{cm}^{h}([\beta_r^- \ ; \ q_{br}^-])-p_{FRI}^{h,d}) $$

\begin{figure}[h!]
	\centering
	\includegraphics[scale=0.3]{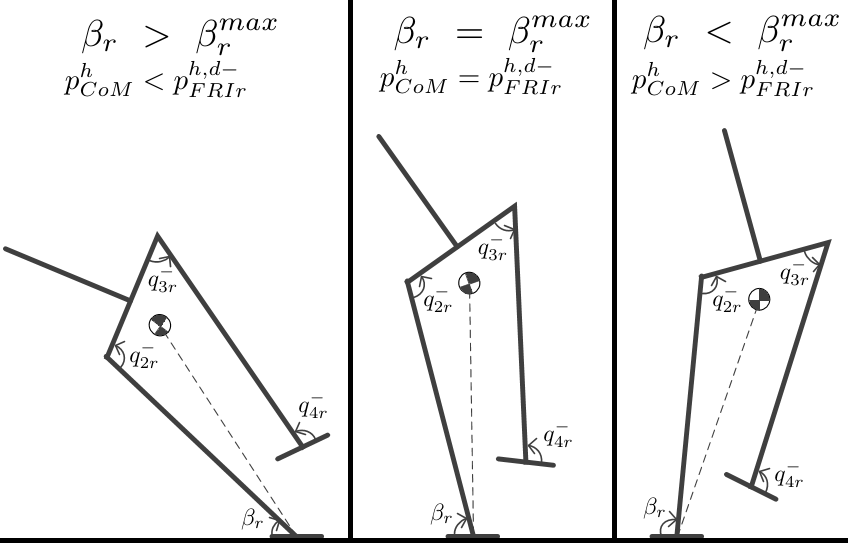}
	\caption{Illustration of different poses and center of mass position of the robot for the same $q_{br}^-$ but for different values of $\beta$ \label{fig:betamax}}
\end{figure}

Notice, that, since the system's pose ($q_{br}^-$) is fixed, the horizontal position of the center of mass is proportional to minus the cosine of $\beta$ --- as illustrated by Fig. \ref{fig:betamax} --- and is monotonically decreasing for $\beta$ in the first two quadrants. Therefore, supposing that $-\pi<\beta_r^-<0$, if $p_{cm}^h([\beta_r^+ \ ; \ q_{br}^-])>p_{FRIr}^{h,d-}$ and $ p_{cm}^h([\beta_r^{max} \ ; \ q_{br}^-])=p_{FRIr}^{h,d-} $, and if $\beta_r^->\beta_r^{max}$, then $ p_{cm}^h([\beta_r^- \ ; \ q_{br}^-])<p_{FRIr}^{h,d-}$ resulting in a positive value of $\dot{\sigma}_{FRI}^-$ contradicting the hypotheses HIP2.

Therefore, if there exists an reversal point it must be in the set $]\beta_r^+, \ \beta_r^{max}[$, a sufficient condition for the existence of this point is that $\zeta_f^*(\beta_r^{max})$ and $\zeta_f^-(\beta_r^{max})$ change signs, this come from the fact that $\zeta_f^*(\beta)$ is continuous in $\beta$.

Find uniqueness conditions for the reversal point, nonetheless, is harder and was not done so far. The authors are currently working on easily verifiable conditions for the uniqueness of the reversal point that will be published in a future work.

\section{Simulations}
\label{sec:simres}

This section presents results for the frontal dynamics control. It is shown that an optimization algorithm can be built and it finds local minimum respecting the specified constraints. The optimization algorithm is built to minimize the functional presented in Eq. (\ref{eq:cost}), where $L_s$ is the step length. Note that this is the usual functional presented in the literature in order to minimize the torque required per distance traveled.%
\begin{equation}
	\label{eq:cost}
	J = \frac{1}{L_s}\int_{0}^{t_{step}}u(t)^\top u(t)dt
\end{equation}

\subsection{Optimization Constraints}

The constraints used in the optimization algorithm were:

\begin{itemize}
	\item[C01] Existence and stability of the limit cycle;
	\item[C02] Existence of the reversal point;
	\item[C03] Invertibility of the decoupling matrix;
	\item[C04] Both ends of the swing foot on the ground during impact;
	\item[C05] Both ends of the swing foot always above ground during step;
	\item[C06] Unilateral constraints satisfied during step;
	\item[C07] Unilateral constraints satisfied at impact;
	\item[C08] Swing leg leaves ground naturally after impact;
	\item[C09] Step duration.
\end{itemize}

\subsection{Optimization Results}

\begin{figure}[h!]
	\centering
	\includegraphics[scale=0.5]{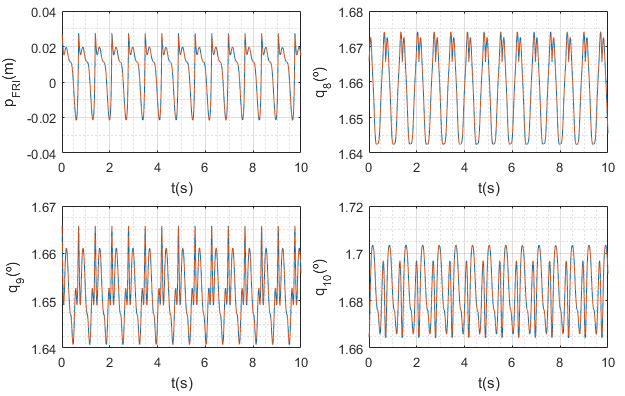}
	\caption{Desired joint value (blue) and simulated joint value (dashed red) \label{fig:FullJointsResults}}
\end{figure}

\begin{figure}[h!]
	\centering
	\includegraphics[scale=0.4]{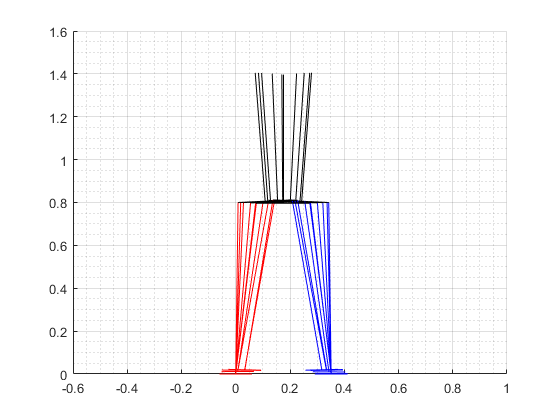}
	\caption{Snapshot of the first two steps of the final optimized frontal gait \label{fig:Snapshot}}
\end{figure}

The initial conditions for the optimization algorithm had a cost of 16268 $N^2m$ and did not respect restrictions C07 and C09. The optimization result respected all nonlinear constraints and resulted in a cost of 4749 $N^2m$, resulting in a decrease of about 80\% of the original cost. The robot's joints desired and simulated values are presented in Fig. \ref{fig:FullJointsResults} and the gait snapshot for the first two steps is presented in Fig. \ref{fig:Snapshot}.

\begin{figure}[h!]
	\centering
	\includegraphics[scale=0.5]{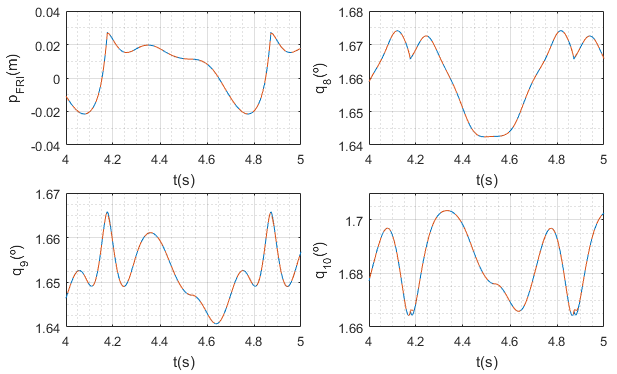}
	\caption{Desired joint value (blue) and simulated joint value (dashed red) \label{fig:ZoomJointsResults}}
\end{figure}

Note that the constraints were satisfied, indicating that the invariance hypothesis is satisfied. It is possible, however, to notice a small time-frame during which $q_{10}$ loses invariance when zooming in as in Fig. \ref{fig:ZoomJointsResults}.

This loss of invariance happens during inversion and is due to imprecisions on the inversion point computation. For comparison, a difference of $10^{-5}$ between the computed and the simulated inversion value resulted on the observed loss of invariance of about $10^{-3}$ radians.

This sensibility of the invariance with the inversion point indicates that any 3D gait designed from this method must deal with the problem of sensitivity to external perturbations on the frontal plane. This is possible to be achieved by changing the stance ankle control law, or by changing the parameters between steps in order to change the foot placement during walking.

\section{Conclusions and Future Works}
\label{sec:conc}

This paper successfully proposes an extension of the hybrid zero dynamics method to control the decoupled frontal dynamics of a bipedal robot and shows solutions for the main difficulties of applying this method to this system.

The two main theoretical points of this paper that need to be better defined are the conditions for the convergence of the successive approximations algorithm used to find the reversal point while assuring hybrid invariance and the conditions under which the reversal point is unique for the given set of parameters.

Even without this, the proposed optimization algorithm still succeeded in finding a local minimum while allowing the insertion of gait style constraints during gait design. 

It also became evident an inherent problem of the frontal dynamics when compared to the sagittal one: sensibility to external perturbations. This problem is equivalent to the one of assuring the existence of the reversal point and can be understood by thinking in terms of potential energy barrier. The sagittal gait must always have more energy than the potential barrier in order to maintain the gait, while the frontal gait must never surpass it or risk overturning during the rising phase.

\addtolength{\textheight}{-12cm}   




\bibliographystyle{plain}
\bibliography{root}{}

\end{document}